\documentclass[11pt,a4paper]{article}
\usepackage[utf8]{inputenc}
\usepackage{times}
\usepackage{comment}
\usepackage{latexsym}
\usepackage{amsmath}
\usepackage{array}
\newcolumntype{P}[1]{>{\centering\arraybackslash}p{#1}}

\usepackage{geometry}
\usepackage[%
	doublespacing, %
	nodisplayskipstretch, 
]{setspace} %

\usepackage[hyperref]{acl2020}

\usepackage{graphicx}
\usepackage{soul}
\usepackage{array}
\usepackage{multicol}
\usepackage{subfig}

\usepackage{mathtools,halloweenmath}
 
\usepackage{lipsum}  
\usepackage{amsmath}

\usepackage{pgfplots}
\pgfplotsset{width=10cm,compat=1.17}

\usepackage{newunicodechar}

\input{titlepage}

\usepackage{arydshln}

\definecolor{mygreen}{RGB}{217, 234, 211}
\definecolor{mygreen2}{RGB}{163, 219, 143}
\definecolor{myred}{RGB}{244, 204, 204}
\definecolor{myred2}{RGB}{244, 100, 100}
\definecolor{mypurple}{RGB}{217, 210, 233}
\definecolor{mypurple2}{RGB}{164, 140, 219}

\usepackage{calligra} 

\input{titlepage} 
\input{lmacros} 

\begin{document}
\title{Few-Shot Cross-Lingual Transfer for
Prompting Large Language Models in
Low-Resource Languages} 
\author{Christopher Toukmaji} 
\degreeyear{2023} 
\degreemonth{June} 
\degree{BACHELOR OF SCIENCE} 
\field{Computer Science} 
\campus{Santa Cruz} 
\chair{Professor Jeffrey Flanigan} 
\committeememberone{Professor Ian Lane} 
\committeemembertwo{Professor Amita Misra} 
\deanlineone{Professor Martine Schlag}
\deanlinetwo{Professor and Undergraduate Director, Computer Science and Engineering}
\deanlinethree{}

\maketitle

\pgfkeys{
  /pgfplots/error bars/error mark and bar options/.code={%
    \pgfkeysalso{
      /pgfplots/error bars/error mark options/.append={,#1},
      /pgfplots/every error bar/.append style={#1}
    }%
  }
}
\pagenumbering{roman}
\setcounter{page}{3}

\tableofcontents

\pagebreak
\section*{}
\addcontentsline{toc}{section}{Figures and Tables}

         \begin{center}
            \Large
            \textbf{Figures and Tables}
         \end{center}
         \normalsize 
\listoffigures
\listoftables
\pagebreak

    \begin{titlepage}
    \begin{center}

        \begin{abstract}
        \begin{centering}
        \vspace{0.35cm}
        Few-Shot Cross-Lingual Transfer for Prompting Large Language Models in Low-Resource Languages
        \vspace{0.35cm}
        \\
        by
        \\
        \vspace{0.35cm}Christopher Toukmaji \\
        \end{centering}
        \vspace{0.35cm}
                
        Large pre-trained language models (PLMs) are at the forefront of advances in Natural Language Processing. One widespread use case of PLMs is ``prompting'' - or  in-context learning  - where a user provides a description of a task and some completed examples of the task to a PLM as context before prompting the PLM to perform the task on a new example. Only the largest, most capable PLMs are able to perform in-context learning effectively, and these models are typically trained with a predominantly English corpus, leaving all other languages behind. The data limitations in most languages preclude the training of language-specific PLMs capable of prompting.
        Albeit the surge in work of prompting settings, it is still unclear how PLMs should be adapted cross-lingually specifically for prompting. We evaluate the possible methods to adapt LLaMa, a 7B parameter open-source PLM mainly trained in English, for prompting in low-resource languages, namely for Kinyarwanda (\texttt{kin}), Hausa (\texttt{hau}), and Luganda (\texttt{lug}). We consider three methods: few-shot prompting (\textbf{\emph{prompt}}), language-adaptive fine-tuning (\textbf{\emph{LAFT}}), and neural machine translation (\textbf{\emph{translate}}), and evaluate on abstractive summarization, multi-class topic classification, and named-entity recognition. Although LAFT carries the greatest compute cost and intuitively should lead to the best results, our experiments exhibit that LAFT is only occasionally the optimal choice for adapting PLMs for prompting. Rather, the translate and prompt settings are a compute-efficient and cost-effective method of few-shot prompting for the selected low-resource languages. We find that the results are task and language dependent but find that the prompting method is the best on average across all tasks and languages. Results show that the prompt setting performs better than both translating and LAFT with statistical significance for all shots when aggregated across all tasks and languages.
        \end{abstract}

    \end{center}

\pagebreak

\section*{}
\setcounter{page}{8}
\addcontentsline{toc}{section}{Dedication}

\begin{center}
\vspace*{\fill}

    \begingroup
  \calligra\huge
	For my parents
\endgroup
\vspace*{\fill}

\end{center}

\pagebreak

    \newpage
    \section*{}
    \addcontentsline{toc}{section}{Acknowledgements}
    \begin{center}
         \vspace*{2cm}
         \Large
         \textbf{Acknowledgements}
         \normalsize \\ 
         \vspace*{1.5cm}
    \end{center}
         I would first like to thank my advisor and the chair of the committee, Dr. Jeffrey Flanigan. Professor Flanigan welcomed me to his lab in my junior year, and he provided unparalleled mentorship during my research journey. I am incredibly grateful to be have been given this early exposure to research in the Natural Language Processing field. This opportunity has allowed me to expand my learning beyond the classroom and has invaluably shaped my future endeavors.

         I am also thankful to the committee members, Dr. Ian Lane and Dr. Amita Misra, for their help and time reviewing my work.  
         
         I would further like to thank the other students in Dr. Jeffrey Flanigan's lab for their support and companionship: Brendan King, Brian Mak, Changmao Li, Chris Liu, Geetanjali Rakshit, Nilay Patel, Rongwen Zhao, and Zekun Zhao.

         I would also like to thank my parents and my friends for their everlasting support.
         
         Lastly, I am thankful for the computing resources provided by the Pacific Research Platform's Nautilus cluster, supported by the National Science Foundation under Award Numbers CNS-1730158, ACI-1540112, ACI1541349, OAC-1826967, the University of California Office of the President, and the University of California San Diego’s California Institute for Telecommunications and Information Technology/Qualcomm Institute.
\end{titlepage}

\pagenumbering{arabic}

\section{Introduction}

        Large pre-trained language models (PLMs) have been at the forefront of the advancements in Natural Language Processing (NLP), evidenced by state-of-the-art results on numerous benchmarks. General purpose PLMs tend to solve many tasks successfully and can reach even better performance on a task when the PLM has been adapted to perform that task.

        One prevalent approach to PLM adaptation is \emph{fine-tuning}, where all of a model's weights are updated given task data. Fine-tuning a PLM on a downstream task may boost the performance for that specific task ~\citep{BERT}, but this improvement comes at the expense of forgetting the learned general knowledge implicitly stored in the model's parameters during pre-training ~\citep{forgetting}. Moreover, traditional fine-tuning is computationally expensive due to the cost for electricity and GPU training time ~\citep{strubell-etal-2019-energy}. Fine-tuning may also fail in regimes with limited training data.
        
        Alternatively, prompting - also known as \emph{in-context learning} - is a substitute to fine-tuning and does not suffer from the common issues of fine-tuning such as forgetting, expensive compute cost, and a lack of training data. Researchers discovered that, with prompting, large PLMs are able to learn downstream tasks after seeing some example completions - or \emph{shots} - of the task within the prompt itself. Prompting can reach comparable results to fine-tuning with a fraction of the data ~\citep{scaorush_worth}. The in-context learning capability of PLMs emerges as the number of parameters in the PLM increases, a phenomenon known as an ``emergent ability'' \cite{wei2022emergent}. Prompting was introduced in GPT-2 in the zero-shot setting \cite{gpt2} and in GPT-3 \cite{gpt3} in the few-shot, or in-context learning, setting. This method requires zero gradient updates.

        Prompting does not require any annotated data, so prompting PLMs can help in settings with limited data. One such setting is low-resource languages - languages that do not have a large amount of annotated data which is needed for fine-tuning. These languages are sometimes widely spoken despite being low-resource. Resource limitations prevent speakers of low-resource languages from participating in modern-day NLP since PLMs need considerable amounts of training data. This exclusion is a particularly crucial issue, as most languages are low-resource, and these languages have billions of speakers \cite{magueresse2020lowresource}.

        \begin{figure}
        \centering
        \includegraphics[scale=0.18]{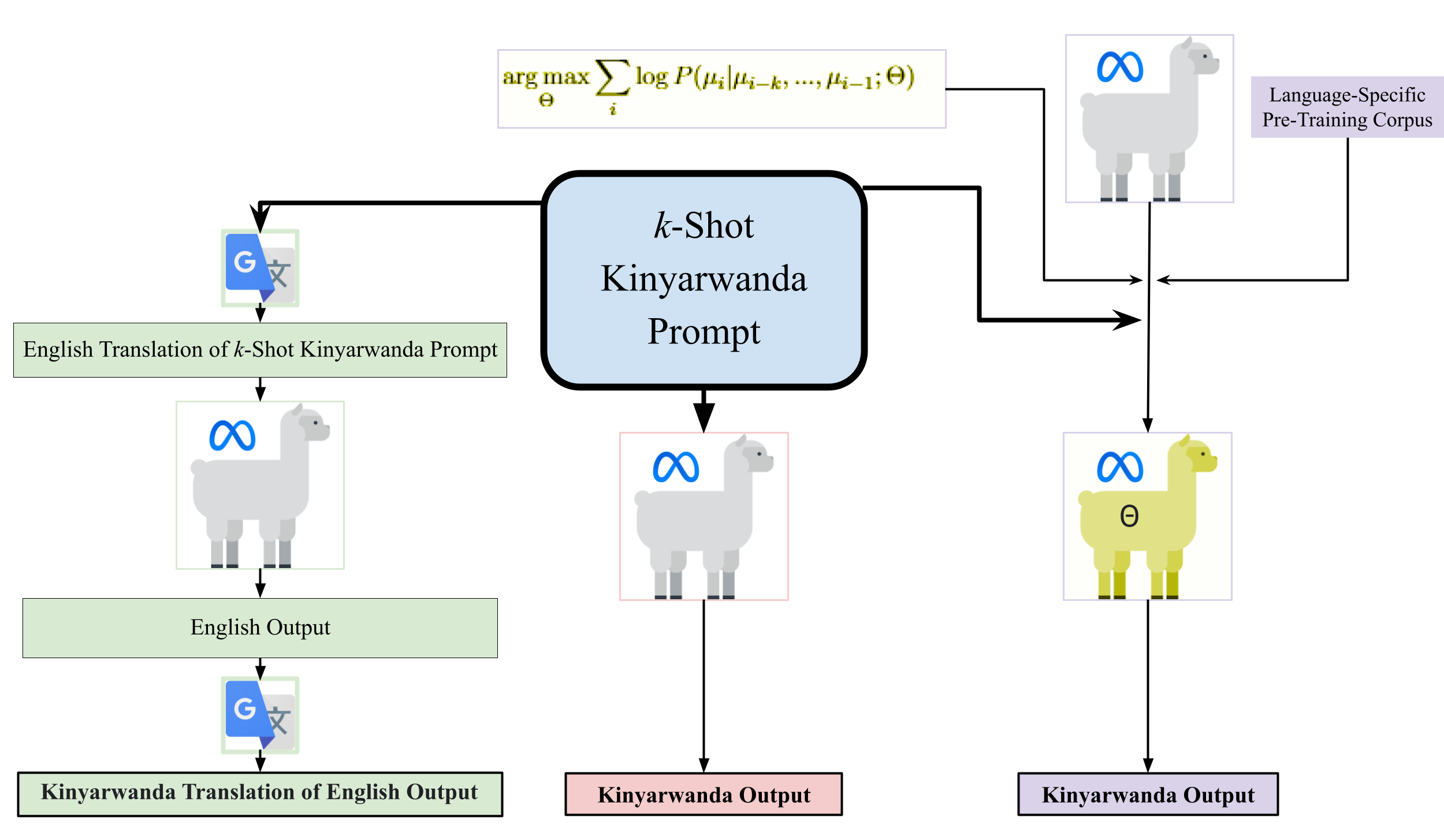}
        \caption{Overview Diagram}
        \caption*{We prompt LLaMa with \emph{k}-shots in a Low-Resource Language in three settings: \colorbox{mygreen}{Translating}, \colorbox{myred}{Prompting}, and \colorbox{mypurple}{language-adaptive fine-tuning}}
        \hspace{0.5in} 
        \noindent\rule{14cm}{0.4pt}
        \end{figure}

        Annotation in low-resource NLP typically involves human annotation or cross-lingual models: using models trained in high-resource source languages for low-resource target languages. Human annotation can be expensive and  - in some cases - implausible, due to a lack of access to fluent speakers in low-resource settings. One common cross-lingual approach involves training models from scratch on several languages  \cite{XLM, mbert, curse-multilinguality, mbart, xue2021mt5,afriberta,XGLM}. However, as more languages are introduced in the training data of the model, the monolingual and cross-lingual performance deteriorates - a phenomenon known as the \emph{curse of multilinguality} \cite{curse-multilinguality}. 

        The approach we employ to help mitigate this problem utilizes high-performance primarily-monolingual PLMs and evaluates them on prompting regimes in low-resource languages. Albeit the recent surge in work analyzing fine-tuning and prompting settings, it is still unclear how PLMs should be adapted cross-lingually specifically for prompting. We evaluate the cross-lingual prompting capabilities of low-resource languages within LLaMa \cite{touvron2023llama}, a 7-billion parameter open-source PLM predominantly trained in English. In this work, we define English as the source language and any non-English language as the target language. Given a PLM that is trained in the source language, we want to adapt this PLM for prompting in the target language. We aim to address the following question:
        \begin{itemize}
            \item \emph{\textbf{Research Question:} Which of the following settings best adapts a large pre-trained monolingual language model from the source language to the target language when prompting in a low-resource language: few-shot prompting, language-adaptive fine-tuning then few-shot prompting, or neural machine translation then few-shot prompting?}
        \end{itemize}

    Although LAFT carries a computational cost much greater than both the translate and prompt settings and intuitively should lead to the best results, our experiments exhibit that LAFT is only occasionally the optimal choice for adapting PLMs for prompting. Rather, the translate and prompt settings are a compute-efficient and cost-effective method of few-shot prompting for the selected subset of low-resource languages. We find that the results are task and language dependent but find that the prompting method is the best on average across all tasks and languages.

\section{Contributions}
     We present a systemic evaluation of methods for adapting a large PLM to a low-resource language for prompting. Specifically, we investigate three methods: 1) few-shot prompting, 2) neural machine translation followed by few-shot prompting, and 3) language-adaptive fine-tuning followed by few-shot prompting. \emph{\underline{L}anguage-\underline{A}daptive \underline{F}ine-\underline{T}uning} (LAFT) \cite{adelani2021masakhaner,alabi-etal-2020-massive} is a method to adapt a PLM from the source language to the target language, in which the pre-training task \eqref{eq:1} is performed on a corpus of tokens in the new language. We perform LAFT on LLaMa - a 7-billion parameter monolingual PLM - for each of the selected low-resource languages: Kinyarwanda (\texttt{kin}), Hausa (\texttt{hau}), and Luganda (\texttt{lug}). 
    We present the first systematic evaluation of these methods of PLM adaptation \emph{specifically} for prompting. Abstractive summarization, named-entity recognition, and multi-class topic classification are applied as downstream tasks. We aggregate the results across tasks and languages which exhibits that few-shot prompting performs better than both translating and LAFT with statistical significance for all shots. Lastly, we release the LAFT LLaMa models in Kinyarwanda,\footnote{\url{https://huggingface.co/ChrisToukmaji/llama_kinyarwanda_LAFT}} Hausa,\footnote{\url{https://huggingface.co/ChrisToukmaji/llama_hausa_LAFT}} and Luganda via the HuggingFace model hub.\footnote{\url{https://huggingface.co/ChrisToukmaji/llama_luganda_LAFT}} 

\section{Related Work}
There has been a great deal of work investigating cross-lingual transfer for PLMs, but there is far fewer work that investigates cross-lingual transfer specifically for prompting with a low-resource language on a large monolingual model. To the extent of our knowledge, this is the first work that investigates cross-lingual transfer of low-resource languages in a large, primarily-monolingual, language model for prompting.  We focus on monolingual models because large, high performance models are typically monolingual due to the curse of multilinguality \cite{curse-multilinguality}.

\newcite{zhao-schutze-2021-discrete} is the first work to investigate cross-lingual transfer for PLMs with prompting. They observe that prompting outperforms fine-tuning in few-shot cross-lingual transfer for a Natural Language Inference (NLI) task using XLM-RoBERTa \cite{curse-multilinguality}. Although this is promising work, there are some shortcomings in the experimental setup. For example, the prompting experiments are evaluated after further training; in the work, 256-shot prompting means that the model was further trained on a 256-shot training set; there are not 256-shots in the prompt. Additionally, the fine-tuning procedure is task-specific, so the work is measuring task-specific/vanilla fine-tuning against prompt-based fine-tuning. The work uses XLM-RoBERTa, a multilingual masked-language-model pre-trained on 100 languages, which is then attached to a linear classifier head and then trained on a few-shot NLI dataset in English to perform an NLI task in other languages. The classified output is then mapped by a verbalizer \cite{schick-schutze-2021-just-verbalizer} to obtain the output in the target language. Other works in this area utilize a similar setup, but this is not the autoregressive-style prompting that we wish to investigate. Our work differs from this work since the downstream task in this work is not an autoregressive gradient-free prompting task. Our work aims to address these shortcomings.

\newcite{xlingfewshot} detail a promising start to cross-lingual transfer for prompting in unseen languages. One of the experiments performs prompt-based fine-tuning and same-language in-context learning for a sentiment analysis task parallel across 12 languages, 8 of which are unseen during the pre-training of the multilingual language model XGLM \cite{XGLM}. They find that in most cases, prompt-based fine-tuning performs better than in-context learning when evaluating both seen and unseen languages. Our work differs from this work since they explicitly use a multilingual model, while we use a monolingual model. More importantly, our fundamental question seeks to evaluate which approach is optimal \emph{specifically} for prompting, while they use prompting as a training strategy.\footnote{Although there are no gradient updates, nor any parameters being updated during prompting, it is still referred to as a training strategy.} 
The evaluation of in-context learning in the work is similar to ours, except we also vary the number of shots since our question is specifically relating to prompting.

\newcite{recyclegpt2} present a method to adapt GPT-2 from English to Dutch and Italian. They first train a new tokenizer using byte-pair-encoding (BPE) for the new language, then retrain only the lexical embeddings by further pre-training with large Dutch and Italian corpora. Finally, they fine-tune the updated GPT-2 model on a downstream task. This work overlaps with our interest in adapting PLMs for non-English languages, but our work differs from this work as this work does not evaluate their method on prompting.

\newcite{lmsarefewshotmultilingCOT} introduce multilinguality in Chain-of-Thought prompting for a parallel mathematical reasoning task \cite{wei2023chainofthought}. The task is evaluated in four settings: (1) few-shot prompting (DIRECT), (2) chain-of-thought prompting with the question, step-by-step example, and answer all in the native language (Native-COT), (3) chain-of-thought prompting with the step-by-step example and answer in English, but the question in the native language (En-COT), and (4) chain-of-thought prompting with the question, step-by-step example, and answer all in English (Translate-En). The results on GPT-3 and PALM \cite{chowdhery2022palm} show that the (Translate-En) setting outperforms the other settings for non-English languages, and either outperforms or performs identically as the other settings in English. Our work differs from this work as this work is primarily concerned with Chain-of-Thought prompting, and it does not compare to a fine-tuned baseline.

\newcite{lmsarefewshotmultiling} is the most similar to our work and show that GPT and T5 \cite{raffel2020exploring-t5} models are able to perform few-shot in-context learning for both English and non-English languages for multi-class classification in settings where the context is in English. It is also shown that these models are able to perform few-shot in-context learning for a non-English language when the given shots are in the same non-English language. The true few-shot prompting experiments in monolingual models outperform prompt-based fine-tuning on multilingual models across all their chosen languages where the context and label are the same language. Our work differs from this work as this work does not evaluate on low-resource languages.

\section{Background}
    \subsection{Neural Language Models}
    The primary objective of Language Models is to assign probabilities to the next token given a prefix of tokens. Early language models were trained on word co-occurrences; modern and more advanced neural language models were developed alongside the emergence of Deep Learning \cite{2003firstneuralLMbengio}. Such models included Recurrent Neural Network (RNN) and Long Short-Term Memory (LSTM) \cite{lstm} language models which were the de facto Deep Learning architecture for language modeling until the development of the transformer model. 

    \subsection{Transformers}
    The neural language model we experiment with in this thesis uses the variant of the transformer model used in LLaMa \cite{touvron2023llama}. The transformer is an encoder-decoder neural network architecture proposed by \newcite{transformer} that revolutionized the NLP space. At a high level, the transformer utilizes the self-attention mechanism which learns how much each word within a sentence is related to another word within the same sentence. Attention precludes the need for recurrence and convolution, opening the doors to parallelization via GPUs \cite{deeplearninggpu}. Many state-of-the-art neural language models such as the GPT series \cite{gpt1} and BERT \cite{BERT} use transformer-based variants such as the transformer decoder and the transformer encoder, respectively. The transformer architecture is gaining popularity in image recognition and other computer vision tasks as well \cite{vision-transformer, vision-survey}.

    \subsection{Pre-training} 
    Pre-training neural language models on a large corpus of text allows them to act as general-purpose language models. Pre-training in PLMs is the process where the parameters of a neural language model are trained with a pre-training objective on a pre-training corpus before fine-tuning on the downstream task.  Popular pre-training objectives include masked-language modeling (MLM) and the ``left-to-right", or autoregressive, language modeling objective. The masked-language modeling objective masks a subset of the words of the words in a sentence, then uses the surrounding context to predict the masked word. On the other hand, the autoregressive language modeling objective trains next-token prediction. These pre-training objectives have led to masked-language models like BERT \cite{BERT}, BERT variants \cite{liu2019roberta}, and autoregressive language models like LLaMa \cite{touvron2023llama} and the GPT series \cite{gpt1, gpt2, gpt3, openai2023gpt4}, respectively.
    
    Given an large untrained randomly-initialized neural language model with parameters $\Theta$,  a context window of size $k$, and an unsupervised corpus of tokens $U$ where
    $U = \{\mu_1, \mu_2, ..., \mu_n\}$,
        the autoregressive pre-training objective maximizes the quantity 
        \begin{equation}\label{eq:1}
        L_1(U) = \sum_{i} \log P(\mu_i | \mu_{i-k}, ..., \mu_{i-1}; \Theta)
        \end{equation}
        following the definition from \newcite{gpt1}.

    \subsection{Language-adaptive fine-tuning} Language-adaptive fine-tuning is the process where a PLM is adapted to a new language by performing the pre-training objective on a corpus of tokens in the new language. LAFT has lead to improved performance in tasks in low-resource languages \cite{adelani2021masakhaner,alabi-etal-2020-massive}.

    \subsection{Fine-tuning} 
    Fine-tuning neural language models for downstream tasks can improve the performance for that specific task. PLMs act as general-purpose language models since they are pre-trained on a large corpus of text. In some cases, however, it may be desirable for a PLM to perform well for a specific task without a need to perform well for other tasks. One method of adapting a PLM for a specific task is fine-tuning, where the weights in a PLM are altered to increase performance for a specific use case. 
    
    Given a pre-trained neural language model with parameters $\Theta$ and a labeled corpus $C$ of size $n$, where $y_i$ is the label for a sequence $x_i$ of length $m$, such that $m = (x_i^{1}, x_i^{2}, ..., x_i^{m})$, fine-tuning maximizes the quantity
            \begin{equation}\label{eq:2}
        L_2(C) = \sum_{i}^{n} \log P(y_i | x_{i}^{1}, x_{i}^{2}, ... ,  x_{i}^{m}; \Theta)
        \end{equation}
    with respect to $\Theta$, following the definition from \newcite{gpt1}.
    
    \subsection{Prompting}
    Prompting is an alternative to fine-tuning that does not suffer from the issues of fine-tuning such as forgetting, expensive compute cost, and a lack of training data.
    In general, the in-context learning ability of a PLM increases alongside the number of examples in the prompt passed to the PLM, as well as the number of parameters in the PLM \cite{gpt3, wei2022emergent}. More advanced prompting methods have been developed such as Chain-of-Thought prompting \cite{wei2023chainofthought} which outlines a step-by-step process within the shots of the prompt and leading to better performance on multi-hop reasoning tasks. 
    
    Another popular branch of prompting learns a continuous representation in a section of the prompt rather than only using discrete tokens. Such approaches include P-Tuning \cite{liu-etal-2022-ptuning}, Prompt-Tuning \cite{lester2021power-prompttuning}, and Prefix-Tuning \cite{li-liang-2021-prefix}.

    \subsection{Multilinguality in PLMs}
    There have been some early methods to make PLMs more multilingual.
    Once such method trains multi-lingual PLMs by pre-training a transformer-based language model from scratch on monolingual corpora consisting of several languages such as XLM \cite{XLM},  mBERT \cite{mbert}, XLM-RoBERTa \cite{curse-multilinguality}, mBART \cite{mbart}, mT5 \cite{xue2021mt5}, AfriBERTa \cite{afriberta}, and XGLM \cite{XGLM}. However, as more languages are introduced, the monolingual and cross-lingual performance deteriorates - a phenomenon known as the \emph{curse of multilinguality} \cite{curse-multilinguality}.  There has been some work in mitigating the curse of multilinguality. Namely, X-Mod is an approach where \cite{pfeiffer2022liftingcurse} language-specific multilayer perceptrons (MLPs) are placed in every transformer layer.

    Another method is language-specific language models, which consists of pre-training a LM from scratch solely for a non-English language such as French \cite{martin-etal-2020-camembert}, Indonesian \cite{koto-etal-2020-indonesian-scratch, wilie-etal-2020-indonesian-scratch2}, Italian \cite{Polignano2019-italianscratch}, and Indian languages/dialects \cite{kakwani-etal-2020-indicnlpsuite-indianLM}.

    Other methods include multilingual adaptive fine-tuning \cite{african-langauge-adaptive-ft}, an extension of language adaptive fine-tuning that pre-trains on many languages, and the method of \newcite{recyclegpt2} in which the lexical embeddings are updated by further pre-training while the rest of the PLM is frozen.

\section{Methods}

        We compare three methods for adapting an PLM trained in a source language for prompting with a target language. We define English as the source language and any non-English language as the target language.
            \begin{enumerate}
        \item \textbf{\emph{Prompt}:} Prompt the PLM in the target language.
        \item \textbf{\emph{Translate}:} Automatically translate the prompt from the target language to the source language, prompt the PLM with the translation, then translate the output back from source language to the target language.
        \item \textbf{\emph{Language-adaptive fine-tuning (LAFT)}:} Perform language-adaptive fine-tuning on the PLM on the target language, 
        then prompt the PLM in the target language.

        \end{enumerate}

        We experiment with three languages (Kinyarwanda, Hausa, and Luganda) on three NLP tasks: Named-Entity Recognition (NER), abstractive summarization, and multi-class topic classification. Not every task is evaluated in every language because we do not have datasets for all these tasks in each language.   
        Each task and language are evaluated independently, then the results are aggregated to observe which specific method performs the best on average.

        \subsection{Method Details}
            In this section, we describe the details for each method. 
            For all settings, each shot is prepended with a machine-translated natural language description of the task in the native language of the prompt.  We prompt the PLM for each element of the test sets outlined in \S \ref{dataset-eval}.  An example few-shot prompt is shown in Figure \ref{fig:fig2}.

            \subsubsection{\emph{Prompt}}
            \label{promptingequations}
            We prompt the PLM with the few-shot prompt in the target language and evaluate the completion. This method requires no translation, nor any gradient updates.
            
            \subsubsection{\emph{Translate}}
            \label{promptingequations2}
            We first translate the few-shot prompt from the target language to the source language using a machine translation system. Next, the PLM is prompted in the source language. Then, the output is translated from the source language back to the target language using a machine translation system. 
            We use the Google Cloud Translate API\footnote{\url{https://cloud.google.com/translate/docs/reference/rest}} for translation, and we provide the estimated costs for translation in \S \ref{reproducibility}.

            \subsubsection{\emph{Language-adaptive fine-tuning}}
            Starting with the original PLM, we further pre-train the PLM on the target language using the original pre-training objective \eqref{eq:1} for a set number of epochs (details given in \S \ref{hyperparams}). Then, we prompt the PLM with the few-shot prompt.

    \begin{figure}\centering

    \noindent\fbox{\begin{minipage}{\textwidth-20\fboxsep-15\fboxrule\relax}

    \textcolor{black}{Shyira buri jambo mu nteruro ikurikira hamwe na tagi ya NER. \\ \\ Ibi ariko uwari Minisitiri w’Ububanyi n’Amahanga w’u Rwanda , Ambasaderi André Bumaya yabiteye utwatsi , abwira PanaPress ikorera i Dakar muri Senegal ko ahubwo ari Uganda yari mu macenga yo gushaka gutera u Rwanda . $\to$ \textbf{(Ibi,O), (ariko,O), (uwari,O), (Minisitiri,O), (w’Ububanyi,O), (n’Amahanga,O), (w’u,O), (Rwanda,B-LOC), (,,O), (Ambasaderi,O), (André,B-PER), (Bumaya,I-PER), (yabiteye,O), (utwatsi,O), (,,O), (abwira,O), (PanaPress,B-ORG), (ikorera,O), (i,O), (Dakar,B-LOC), (muri,O), (Senegal,B-LOC), (ko,O), (ahubwo,O), (ari,O), (Uganda,B-LOC), (yari,O), (mu,O), (macenga,O), (yo,O), (gushaka,O), (gutera,O), (u,B-LOC), (Rwanda,I-LOC), (.,O)} \\ \\ 
    Shyira buri jambo mu nteruro ikurikira hamwe na tagi ya NER. \\ \\ Ati “ Hari imishinga y’imihanda , iy’amazi , iy’amashanyarazi , kugira ngo amafaranga agere ku bantu """" . $\to$ \textbf{(Ati,O), (“,O), (Hari,O), (imishinga,O), (y’imihanda,O), (,,O), (iy’amazi,O), (,,O), (iy’amashanyarazi,O), (,,O), (kugira,O), (ngo,O), (amafaranga,O), (agere,O), (ku,O), (bantu,O), ("""",O), (.,O)} \\  \\}\textit{Shyira buri jambo mu nteruro ikurikira hamwe na tagi ya NER. \\ \\ Bazwi mu cyo bise ‘Morning Worship’ aho baririmba ibihangano bitandukanye byo mu gitabo bigafasha benshi . $\to$}
    \end{minipage}}
    \caption{A 2-shot prompt example}
    \caption*{Two examples for NER in Kinyarwanda with \textcolor{black}{sampled from the train set} (top), \textbf{\textcolor{black}{the respective labels}} and followed by an \textit{input instance of the test set (bottom).}}
    \hspace{0.5in}     
    \noindent\rule{14cm}{0.4pt}
    \label{fig:fig2}
    \end{figure}

\section{Experimental Details}
        We use LLaMa \cite{touvron2023llama}, a 7-billion parameter monolingual PLM, for prompting in three low-resource languages: Kinyarwanda, Hausa, and Luganda. The downstream tasks we evaluate on are abstractive summarization, named-entity recognition, and multi-class topic classification. 

        During evaluation, we form our few-shot prompts with a random sample of $k \in \{1, 2, 4\}$ shots from a language's training split on the evaluation datasets without replacement. We report results on the test split of the dataset for that language. We conduct 5 samples and average the performance across the test split.

        \subsection{Model}
        We use the 7B parameter LLaMa model since it behaves similarly to GPT-3 (see \S \ref{gpt3llama}),
        and it is light-weight and open-source. Moreover, evaluations of LLaMa show that the model is capable of in-context learning which further motivates the use for LLaMa. Due to the simplicity of the selected tasks, we limit the amount of generated tokens to 256. We report the remaining model hyperparameters in $\S$ \ref{hyperparams}. We use PyTorch \cite{paszke2019pytorch}, and the HuggingFace \texttt{transformers} \cite{wolf2020huggingfaces} and \texttt{datasets} \cite{lhoest2021datasets} libraries for efficient model implementation.

         For LAFT, due to the expensive overhead cost of performing hyperparameter optimization, we re-use the hyperparameters used to fine-tune LLaMa 7B for the Alpaca 7B model \cite{alpaca}. While Alpaca was fine-tuned with a sequence length of 512, we used the maximum sequence length of LLaMa, 2048. The remainder of the hyperparameters are identical to those that were used to fine-tune Alpaca.

        \subsection{Languages}
        We select a subset of low-resource African languages \cite{adelani2021masakhaner}: Hausa (\texttt{hau}), Kinyarwanda (\texttt{kin}),  Luganda (\texttt{lug}). The training corpus of LLaMa is predominantly in English, with the exception of a set of Wikipedia pages in 20 languages. The Wikipedia pages make up less than 5\% of the total training data and only include languages of Latin or Cyrillic scripts. We selected the above three non-English languages as they are low-resource, typographically-written with Latin script for the most part, and not well-represented in the LLaMa training set.

        \subsection{Datasets}

            \subsubsection{Language-adaptive fine-tuning}
            
            We use the following language specific fine-tuning corpora for LAFT.  For Hausa, we use the Hausa split of mC4 \cite{xue2021mt5}, a multilingual variant of the C4 dataset \cite{raffel2020exploringc4}. For Kinyarwanda and Luganda, we use CommonVoice \cite{commonvoice_ardila2020}.
            \label{dataset-fine-tune}

            The Hausa dataset is 1.16 GB which consists of 252,446,724 tokens. The Kinyarwanda dataset is 52.96 MB which consists of 11,687,639 tokens. The Luganda dataset is 0.145 MB which consists of 33,184 tokens.

            \subsubsection{Evaluation}
            \label{dataset-eval}
            We evaluate with abstractive summarization, multi-class topic classification, and named-entity recognition as downstream tasks; the selected evaluation datasets are MasakhaNER \cite{adelani2021masakhaner} , XL-Sum \cite{hasan-etal-2021-xlsum}, and KinNEWS \cite{niyongabo2020kinnews}. In order to select evaluation datasets, we utilized the Multilingual Dataset Survey Database \cite{multilingdatasetsurvey} by filtering for datasets with gold labels that were not machine-translated for each of our selected languages.

            For Hausa, we evaluate on MasakhaNER
            and XL-Sum.
            For Kinyarwanda, we evaluate on MasakhaNER
            and KinNEWS.
            For Luganda, we evaluate on MasakhaNER.

            MasakhaNER is a Named-Entity Recognition sequence-labelling task where each space-separated word in a sentence is assigned exactly one of the following tags: O (Non-entity), PER (Personal Name), ORG (Organization), LOC (Location), DATE (Date \& Time). When evaluating MasakhaNER, we concatenate the original sentence with the NER tags for every space-separated word in the sentence. These are the same gold labels provided in the dataset. 
            
            KinNEWS is a task that classifies news headlines into 14 distinct classes represented by integers.
            When evaluating KinNEWS, we form the prompts by concatenating the title of each news headline with the provided integer label corresponding to the topic of the news article. From the languages contained within KinNEWS, only Kinyarwanda is in subset of selected low-resource languages.
            
            XL-Sum is an abstractive summarization task of news articles. When evaluating XL-Sum, we concatenate the news article with the provided summary of the news article. We provide sample prompts for all experiments in $\S$ \ref{Appendix}. From the languages contained within XL-Sum, only Hausa is in the subset of selected low-resource languages.
            
            We provide a majority-label baseline for KinNEWS - the only one of the selected evaluation tasks where a majority-label baseline can be applied.

         \subsection{Metrics}
            We use F1-score, accuracy, and ROUGE score \cite{lin-2004-rouge} as evaluation metrics. 
            The evaluation metrics depend on the downstream task. 
            
            MasakhaNER is a multi-class classification task, so we report the F1-score following their work. F1-score can be written in terms of precision and recall. We use the CoNLL-2003 definition of precision and recall for name-entity recognition which states that ``Precision is the percentage of named entities found by the learning system that are correct. Recall is the percentage of named entities present in the corpus that are found by the system. A named entity is correct only if it is an exact match of the corresponding entity in the data file". \cite{tjong-kim-sang-de-meulder-2003-introduction-conll}

            KinNEWS is a multi-class classification task. The paper reports accuracy scores, so we will report accuracy. We clean the generated sequences by using the first instance of a space-separated integer in the output as the predicted label. 
            
            XL-Sum is a summarization task, so we report ROUGE-1, ROUGE-2, and ROUGE-L, following the work. 

    \begin{table*}[h]\centering
    \begin{tabular}{c|c||P{3cm} P{3cm} P{3cm}}
    \hline \hline Dataset and (Metric) &\textbf{Method} & \emph{Prompt} & \emph{Translate} & \emph{LAFT} \\ \hline
    & 1-shot  & \textbf{0.1245} & 0.0341 & 0.0723 \\
    MasakhaNER & 2-shot  & \textbf{0.2880} & 0.0906 & 0.1083 \\
    (F1-Score) & 4-shot  & \textbf{0.3507} & 0.1558 & 0.1737 \\
    \hline
    & 1-shot  & (0.1421 / \textbf{0.0400} / 0.0985) &  (0.0267 / 0.0044 / 0.0239) & (\textbf{0.1478} / 0.0373 / \textbf{0.1050}) \\
    XL-Sum  & 2-shot  & (0.1193  / 0.0295 / 0.0847) &  (0.0260 / 0.0046 / 0.0233) 
 & (\textbf{0.1445} / \textbf{0.0373} / \textbf{0.1049}) \\
    (ROUGE-1/-2/-L)& 4-shot  & (0.1195 / 0.0290 / 0.0846) &  (0.0268 / 0.0050 / 0.0241) 
 & (\textbf{0.1419} / \textbf{0.0346} / \textbf{0.1040}) \\
    \hline
    \hline
    \end{tabular}
    \caption[Performance on the Hausa (\texttt{hau}) test split for each dataset.]{\label{demo-table1} Performance on the Hausa (\texttt{hau}) test split for each dataset. Results averaged for 5 seeds. For MasakhaNER, we report F1-Score.  For XL-Sum, we report ROUGE-1/ROUGE-2/ROUGE-L.}
    \end{table*}

    \begin{table*}[h]
    \centering
    \begin{tabular}{c|c||ccc}
    \hline \hline Dataset and (Metric) & \textbf{Method} & \emph{Prompt} & \emph{Translate} & \emph{LAFT} \\ \hline
    & 1-shot  & 0.0133 & \textbf{0.0281} & 0.0000 \\
    MasakhaNER & 2-shot  & \textbf{0.1643} & 0.1144 & 0.0000 \\
    (F1-Score) & 4-shot  & 0.1432 & \textbf{0.1705} & 0.0011 \\
    \hline
    \hline

    & 1-shot  & 0.1156 & \textbf{0.1175} & 0.0123 \\
    KinNEWS  & 2-shot  & 0.1746 & \textbf{0.1910} & 0.1686 \\
    (Accuracy) & 4-shot  & 0.1726 & \textbf{0.2006} & 0.1547 \\
    & Majority Baseline & 0.2301 & 0.2301 & 0.2301 \\
        
    \hline
    \hline
    \end{tabular}
    \caption[ Performance on the Kinyarwanda (\texttt{kin}) test split for each dataset.]{\label{demo-table2} Performance on the Kinyarwanda (\texttt{kin}) test split for each dataset. Results averaged for 5 seeds. For MasakhaNER, we report F1-Score. For KinNEWS, we report Accuracy.}
    \end{table*}

    \begin{table*}[h]
    \centering
    \begin{tabular}{c|c||ccc}
    \hline \hline Dataset and (Metric) & \textbf{Method} & \emph{Prompt} & \emph{Translate} & \emph{LAFT}\\ \hline
    & 1  & 0.0664 & \textbf{0.0780} & 0.0316 \\
    MasakhaNER & 2  & \textbf{0.1312} & 0.0857 & 0.0670 \\
    (F1-Score) & 4  & \textbf{0.1601} & 0.1435 & 0.0975 \\
    \hline
    
    \hline
    \hline
    \end{tabular}
    \caption[Performance on the Luganda (\texttt{lug}) test split for each dataset.]{\label{demo-table3} Performance on the Luganda (\texttt{lug}) test split for each dataset. Results averaged for 5 seeds. For MasakhaNER, we report F1-Score.}
    \end{table*}

\section{Results}
    \setcounter{figure}{2}    
    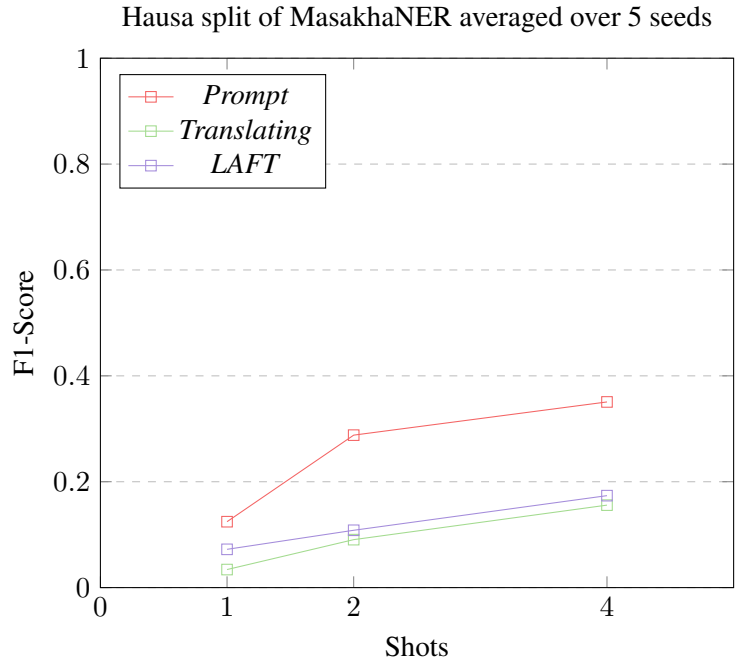
\begin{figure}
    \centering
    \begin{tikzpicture}
    \begin{axis}[
        title={Hausa split of MasakhaNER averaged over 5 seeds},
        xlabel={Shots},
        ylabel={F1-Score},
        xmin=0, xmax=5,
        ymin=0, ymax=1,
        xtick={0,1,2,4},
        ytick={0,0.2, 0.4, 0.6, 0.8, 1},
        legend pos=north west,
        ymajorgrids=true,
        grid style=dashed,
    ]
    \addplot[
        color=myred2,
        mark=square,
        ]
        coordinates {
        (1,0.1245)(2,0.2880)(4, 0.3507)
         };
        \legend{\emph{Prompt}}
    
    \addplot[
        color=mygreen2,
        mark=square,
        ]
        coordinates {
        (1,0.0341)(2,0.0906)(4, 0.1558)
         };
        \addlegendentry{\emph{Translating}}

    \addplot[
        color=mypurple2,
        mark=square,
        ]
        coordinates {
        (1,0.0723)(2,0.1083)(4, 0.1737)
         };
        \addlegendentry{\emph{LAFT}}
        
    \end{axis}
    \end{tikzpicture}
         \caption{\label{graph1} Hausa split of MasakhaNER averaged over 5 seeds}
    \end{figure}

    \begin{figure}
    \centering
    \begin{tikzpicture}
    \begin{axis}[
        title={Hausa split of XL-Sum averaged over 5 seeds},
        xlabel={Shots},
        ylabel={ROUGE-1},
        xmin=0, xmax=5,
        ymin=0, ymax=1,
        xtick={0,1,2,4},
        ytick={0,0.2, 0.4, 0.6, 0.8, 1},
        legend pos=north west,
        ymajorgrids=true,
        grid style=dashed,
    ]
    \addplot[
        color=myred2,
        mark=square,
        ]
        coordinates {
        (1,0.1421)(2,0.1193)(4, 0.1195)
         };
        \legend{\emph{Prompt}}
    
    \addplot[
        color=mygreen2,
        mark=square,
        ]
        coordinates {
        (1,0.0267)(2,0.0260)(4, 0.0268)
         };
        \addlegendentry{\emph{Translating}}

    \addplot[
        color=mypurple2,
        mark=square,
        ]
        coordinates {
        (1,0.1478)(2,0.1445)(4, 0.1419)
         };
        \addlegendentry{\emph{LAFT}}
        
    \end{axis}
    \end{tikzpicture}
             \caption{\label{graph2} Hausa split of XL-Sum averaged over 5 seeds}
    \end{figure}
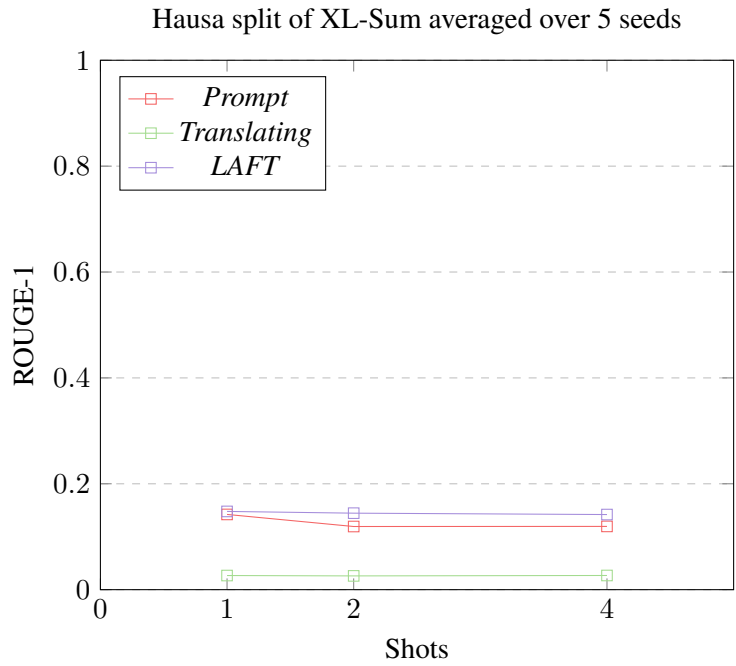

    \begin{figure}
    \centering
    \begin{tikzpicture}
    \begin{axis}[
        title={Kinyarwanda split of MasakhaNER averaged over 5 seeds},
        xlabel={Shots},
        ylabel={F1-Score},
        xmin=0, xmax=5,
        ymin=0, ymax=1,
        xtick={0,1,2,4},
        ytick={0,0.2, 0.4, 0.6, 0.8, 1},
        legend pos=north west,
        ymajorgrids=true,
        grid style=dashed,
    ]
    \addplot[
        color=myred2,
        mark=square,
        ]
        coordinates {
        (1,0.0133)(2,0.1643)(4, 0.1432)
         };
        \legend{\emph{Prompt}}
    
    \addplot[
        color=mygreen2,
        mark=square,
        ]
        coordinates {
        (1,0.0281)(2,0.1144)(4, 0.1705)
         };
        \addlegendentry{\emph{Translating}}

    \addplot[
        color=mypurple2,
        mark=square,
        ]
        coordinates {
        (1,0.0000)(2,0.0000)(4, 0.0011)
         };
        \addlegendentry{\emph{LAFT}}
        
    \end{axis}
    \end{tikzpicture}
    \caption{\label{graph3} Kinyarwanda split of MasakhaNER averaged over 5 seeds}
    \end{figure}
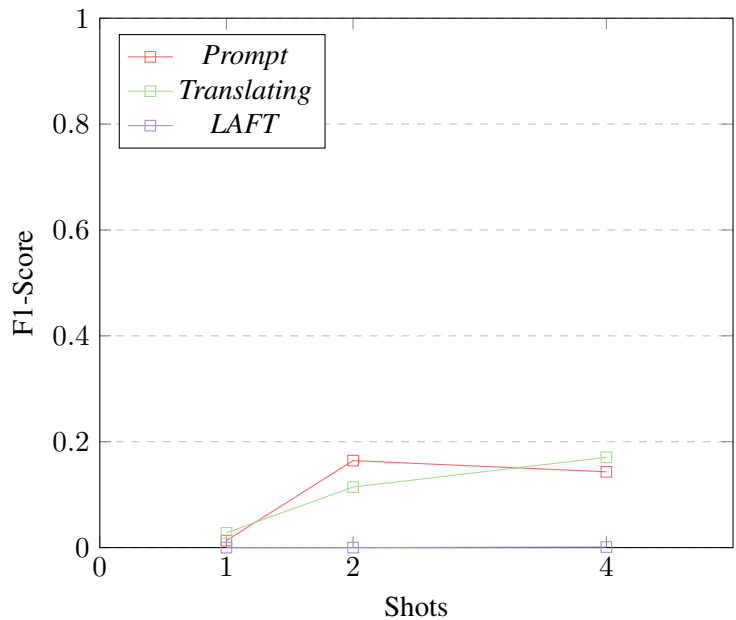

    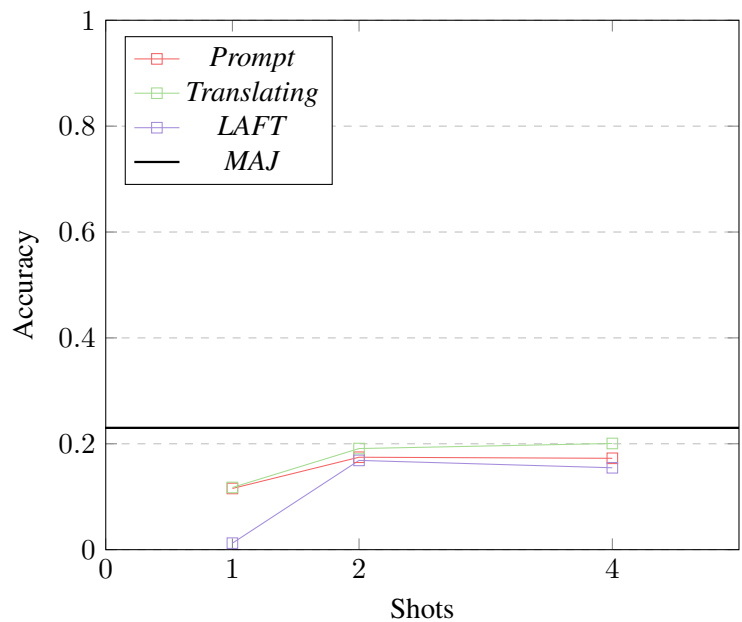
\begin{figure}
    \centering
    \begin{tikzpicture}
    \begin{axis}[
        title={Kinyarwanda split of KinNEWS averaged over 5 seeds},
        xlabel={Shots},
        ylabel={Accuracy},
        xmin=0, xmax=5,
        ymin=0, ymax=1,
        xtick={0,1,2,4},
        ytick={0,0.2, 0.4, 0.6, 0.8, 1},
        legend pos=north west,
        ymajorgrids=true,
        grid style=dashed,
    ]
    \addplot[
        color=myred2,
        mark=square,
        ]
        coordinates {
        (1,0.1156)(2,0.1746)(4, 0.1726)
         };
        \legend{\emph{Prompt}}
    
    \addplot[
        color=mygreen2,
        mark=square,
        ]
        coordinates {
        (1,0.1175)(2,0.1910)(4, 0.2006)
         };
        \addlegendentry{\emph{Translating}}

    \addplot[
        color=mypurple2,
        mark=square,
        ]
        coordinates {
        (1,0.0123)(2,0.1686)(4, 0.1547)
         };
        \addlegendentry{\emph{LAFT}}
    
    \addplot[
        color=black,
        line width=0.3mm
        ]
        coordinates {(-1, 0.2301)(10, 0.2301)};
        \addlegendentry{\emph{MAJ}}
        
    \end{axis}
    \end{tikzpicture}
    \caption{\label{graph4} Kinyarwanda split of KinNEWS averaged over 5 seeds}
    \end{figure}

    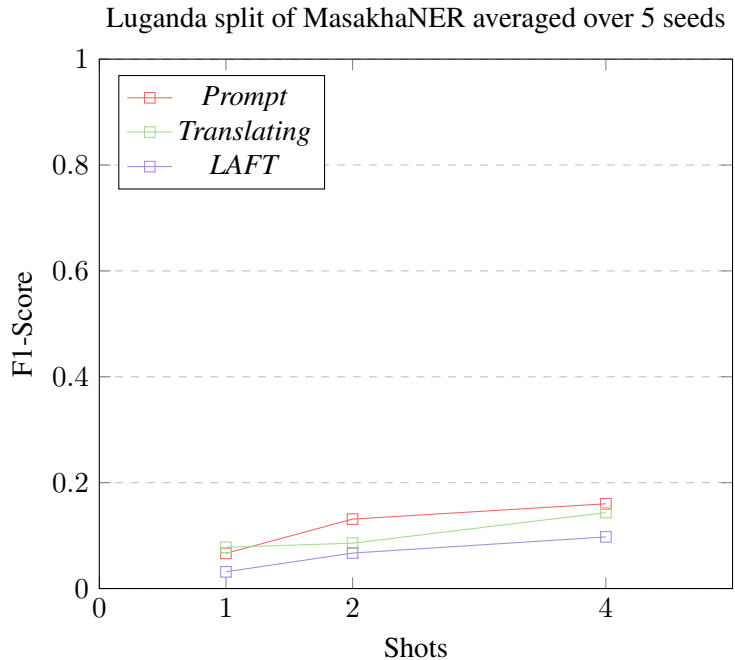
\begin{figure}
    \centering
    \begin{tikzpicture}
    \begin{axis}[
        title={Luganda split of MasakhaNER averaged over 5 seeds},
        xlabel={Shots},
        ylabel={F1-Score},
        xmin=0, xmax=5,
        ymin=0, ymax=1,
        xtick={0,1,2,4},
        ytick={0,0.2, 0.4, 0.6, 0.8, 1},
        legend pos=north west,
        ymajorgrids=true,
        grid style=dashed,
    ]
    \addplot[
        color=myred2,
        mark=square,
        ]
        coordinates {
        (1,0.0664)(2,0.1312)(4, 0.1601)
         };
        \legend{\emph{Prompt}}
    
    \addplot[
        color=mygreen2,
        mark=square,
        ]
        coordinates {
        (1,0.0780)(2,0.0857)(4, 0.1435)
         };
        \addlegendentry{\emph{Translating}}

    \addplot[
        color=mypurple2,
        mark=square,
        ]
        coordinates {
        (1,0.0316)(2,0.0670)(4, 0.0975)
         };
        \addlegendentry{\emph{LAFT}}
        
    \end{axis}
    \end{tikzpicture}
    \caption{\label{graph5} Luganda split of MasakhaNER averaged over 5 seeds}
    \end{figure}

    \begin{table*}[h]
    \centering
    \begin{tabular}{c|c||c}
    \hline \hline Shot & Method & $\bar{x} \pm s $ \\ \hline
    1 & Prompt & $0.0858 \pm 0.0145$ \\
    1 & Translate & $0.0447 \pm 0.0192$ \\
    1 & LAFT & $0.0580 \pm 0.0055$ \\
    \hline
    2 & Prompt & $0.1416 \pm 0.0238 $ \\
    2 & Translate & $0.0765 \pm 0.0205$ \\
    2 & LAFT & $0.0901 \pm 0.0062$ \\
    \hline
    4 & Prompt & $0.1514 \pm 0.0303$ \\
    4 & Translate & $0.1038 \pm 0.0240$ \\
    4 & LAFT & $0.1011 \pm 0.0182$ \\
    \hline
    
    \hline
    \hline
    \end{tabular}
    \caption{Mean and Standard Deviation for Error Bars}
    \end{table*}

    \begin{table*}[h]
    \centering
    \begin{tabular}{c|c|c||c}
    \hline \hline Shot & Group1 & Group2 & p-value \\ \hline
    1&Prompt&Translate&0.0091 \\
    1&Prompt&LAFT&0.0072\\
    1&Translate&LAFT&0.2173\\
    \hline
    2&Prompt&Translate&0.0032 \\
    2&Prompt&LAFT&0.0030\\
    2&Translate&LAFT&0.2410\\
    \hline
    4&Prompt&Translate&0.0391 \\
    4&Prompt&LAFT&0.0215\\
    4&Translate&LAFT&0.8624\\
    \hline
    
    \hline
    \hline
    \end{tabular}
    \caption{p-values after a two-sided t-test for each pair of outputs}
    \end{table*}

We find that the results are task and language dependent, but find that the prompting method is best on average across all tasks and languages. For the Hausa experiments, the prompting setting performed the best in the sequence-labeling task, while language-adaptive fine-tuning performed the best in most cases in the summarization task. For the Kinyarwanda experiments, the translation setting performed the best in the topic classification task, but the best performing method for the sequence-labeling task is dependent on the number of shots in the prompt. The best performing method in Luganda for sequence-labeling was also dependent on the number of shots in the prompt.

We observe that performance increases alongside the number of shots in the prompt in the sequence-labeling task and the news topic classification task, while the performance degrades or stays about the same in summarization. Further analysis of the experimental results expose the settings where prompting is not a viable solution, such as the XL-Sum summarization task in Hausa. Even when a single shot is provided to the PLM, there are some instances of the test set being longer than the context window of 2048 tokens. If the prompt is longer than 2048 tokens, our model truncates the beginning of prompt and only keeps the last 2048 tokens. The proportion of instances that do not fit within the context window increases as the number of shots increase. In turn, this increases the chances of truncation, which leads to the PLM conditioning on a cut-off shot. This is reflected in Table 1 where the ROUGE scores decrease as the number of shots increase. We observe that prompts longer than 2048 tokens occurred across the downstream tasks, but it is much more prevalent and apparent in summarization. 

One surprising result is the strong performance of LAFT on abstractive summarization (XL-Sum), despite LAFT performing poorly on the remainder of the tasks. After manually checking model outputs, we observe that the summarized outputs generated by LAFT contain factually-accurate details about the article that were not mentioned within the article itself. We provide an example of one of these outputs in the appendix (Figure \ref{figure17}). Since the same prompt was used in the prompt experiment (Figure \ref{figure11}) and translate experiment (Figure \ref{figure14}) as well, but did not result in the ``pyschic'' output seen in the LAFT experiment, we suspect there may be data contamination within the Hausa fine-tuning dataset used to train LAFT. 

For sequence labeling tasks like NER, it is not trivial to perform the translation experiments; it is unclear if and how the labels should be translated. In our experiments, we translate the concatenation of the prompt and the labels, prompt, then translate back and NER tags do not have a translation.

    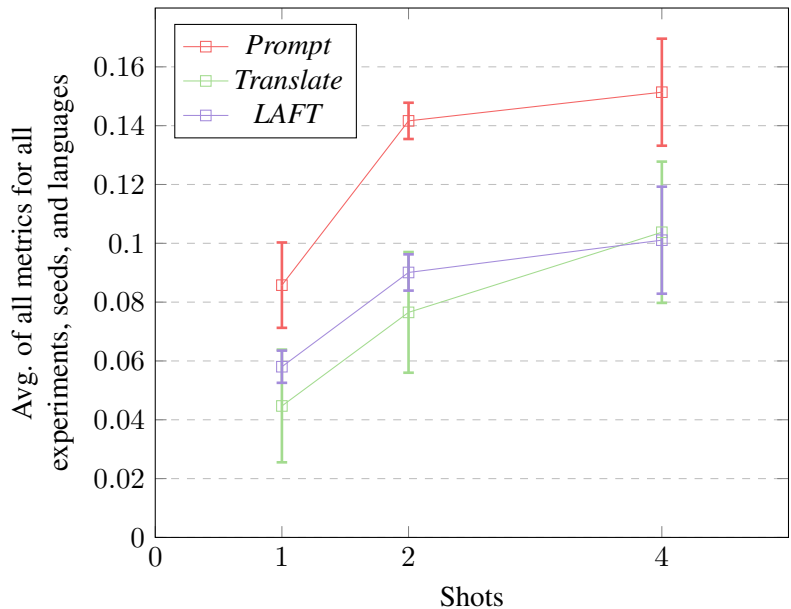
\begin{figure}
    \centering
        \begin{tikzpicture}
        \begin{axis}[
            title={ Aggregated Performance across all tasks for all experiments for all languages},
            xlabel={Shots},
            ylabel style={align=center},
            ylabel={Avg. of all metrics for all\\experiments, seeds, and languages},
            xmin=0, xmax=5,
            ymin=0, ymax=0.18,
            xtick={0,1,2,4},
            ytick={0, 0.02, 0.04, 0.06, 0.08, 0.10, 0.12, 0.14, 0.16},
            legend pos=north west,
            ymajorgrids=true,
            grid style=dashed,
            yticklabel style={/pgf/number format/fixed},
        ]
            \addplot+ [
                color=myred2,
                mark=square,
                error bars/.cd,
                    y dir=both,y explicit,
                    error mark and bar options={line width=1pt, myred2}
            ] table [x=x,y=y,y error=error] {
                x       y       error
                1      0.08577203288853355      0.014511057198365282
                2      0.14163008288926654      0.006166717867881886
                4     0.1513809366301765     0.018194001805781783
            };
                    \addlegendentry{\emph{Prompt}}

            \addplot+ [
                color=mygreen2,
                mark=square,
                error bars/.cd,
                    y dir=both,y explicit,
                    error mark and bar options={line width=1pt, mygreen2}
            ] table [x=x,y=y,y error=error] {
                x       y       error
                1      0.044693639602876034      0.019161698578134358
                2      0.07651816541204645      0.02051757044110845
                4     0.10375764827294245     0.02401763870428815
            };
                    \addlegendentry{\emph{Translate}}

            \addplot+ [
                color=mypurple2,
                mark=square,
                error bars/.cd,
                    y dir=both,y explicit,
                    error mark and bar options={line width=1pt, mypurple2}
            ] table [x=x,y=y,y error=error] {
                x       y       error
                1      0.0580387288434349      0.005474521027088121
                2      0.09008365661567677      0.006166717867881886
                4     0.1010615690536228     0.018194001805781783
            };
                    \addlegendentry{\emph{LAFT}}

        \end{axis}
        \end{tikzpicture}
    \caption{\label{graph6}  Aggregated Performance across all tasks for all experiments for all languages}
    \end{figure}

The only downstream task where a majority baseline could be measured was the multi-class news topic classification in Kinyarwanda, where all the experiments were strictly below the majority baseline, as shown in Table \ref{demo-table2}. Although the translation method is the best setting for this classification task, it still underperforms a weak baseline, so there is still much work to be done within this area. 

Aggregated results show that pure few-shot prompting performs better than both Translating and LAFT with statistical significance $(\alpha = 0.05)$. We provide the results for each of the experiments within Table \ref{demo-table1}, Table \ref{demo-table2}, and Table \ref{demo-table3} and graphical representations. We aggregate Figures 3-7 into Figure \ref{graph6} that measures the average performance of all the tasks for all the languages for every shot and seed. Then, we perform a two-sided t-test on the averages performance for each seed for every pair of shots and methods. Basic few-shot prompting outperforms Translating and LAFT with statistical significance for all shots.

\paragraph{Summary} Our intuition behind why prompting works the best amongst the three settings is shaped by manual examination of model outputs, and we conjecture that prompting works the best due to the following considerations. First, Neural Machine Translation (NMT) systems also suffer from a lack of training data, so the forward and backwards translations are likely noisy. This is particularly observable in the contrast between the performance in unstructured generation tasks, like summarization, and stringently-structured generation tasks, such as NER and topic classification. The downstream task performance of NER and topic classification in translate settings is on par with the other settings, but clearly underperforms in summarization. We suspect this is attributed to the NMT system being more likely to introduce noise into translation of large paragraphs than into translation of NER sequences and integers. Second, LAFT alters all parameters in the PLM, so we suspect that training LAFT degenerates the PLM's capability of instruction-following. We observe that in almost all cases of the MasakhaNER LAFT experiments, the PLM is unable to output the generated text in the structure described in the prompt. Overall, we suspect prompting performs the best as prompting is able to maintain the instruction-following capabilities of PLMs without introducing any noise in translation.

\section{Conclusion}
To the extent of our knowledge, this is the first work that investigates cross-lingual transfer of low-resource languages in a large, primarily-monolingual, language model for prompting. We present an evaluation to measure whether few-shot prompting, language-adaptive fine-tuning then few-shot prompting, or translating then few-shot prompting results in the best performance for prompting in low-resource languages. Although LAFT carries a computational cost much greater than both the translate and prompt settings and intuitively should lead to the best results, our experiments exhibit that LAFT is only occasionally the optimal choice for adapting PLMs for prompting. Rather, the translate and prompt settings are a compute-efficient and cost-effective method of few-shot prompting for the selected subset of low-resource languages. We find that the results are task and language dependent but find that the prompting method is the best on average across all tasks and languages. Aggregated results show that few-shot prompting performs better than both translating and LAFT with statistical significance for all shots. 

\subsection*{Limitations}
\textbf{Data:} We translated with low-resource languages, so we were limited in the amount of pre-training corpora and evaluation data. As a result, we were only able to evaluate on a subset of NLP downstream tasks. \\
\textbf{Machine Translation:} The translation experiments are passed through a neural machine translation system twice, so if the system does not provide high-quality translations, this may negatively bias the results. Future work can investigate the relationship between the performance of the translation approach and the translation quality measured by a Machine-Translation metric such as BLEU or Translation Error Rate. However, the selected evaluation datasets do not have reference translations, so translation quality cannot be evaluated in this specific setting. \\
\textbf{Compute:} The expensive overhead of fine-tuning precludes us from performing hyperparameter optimization. Thus, it is possible that the hyperparameters we selected are not optimal, and we may have a more robust LAFT model after performing hyperparameter optimization. In order to enable a feasible hyperparameter search, parameter efficient fine-tuning methods such as Low-Rank Adapation \cite{hu2021lora} and Quantitization \cite{dettmers2023qlora} have shown the capability to fine-tune multi-billion parameter language models with little to no performance degradation. 

\subsection*{Future Work}
In our work, we experiment with three low-resource languages: Kinyarwanda, Hausa, and Luganda. However, there are thousands of other low-resource languages that are also in need of more research. We hope our work will inspire research for other low-resource languages. 

\bibliography{anthology}

\newpage
\section*{Appendix}
\label{Appendix}
    \
    \subsection{Examples for \emph{Prompt} Experiments}
        \begin{figure}[h]
        \noindent\fbox{\begin{minipage}{\dimexpr\textwidth-80\fboxrule\relax}
        \small
Buli kigambo mu sentensi eno wammanga giteekeko akabonero kaakyo aka NER.
Olina okutegeka byonoobikkisa ngessanja lye lisinga okukola obulungi naye bwoba tolirina ofuna essubi nalyo likola . $\to$ (Olina,O), (okutegeka,O), (byonoobikkisa,O), (ngessanja,O), (lye,O), (lisinga,O), (okukola,O), (obulungi,O), (naye,O), (bwoba,O), (tolirina,O), (ofuna,O), (essubi,O), (nalyo,O), (likola,O), (.,O)
\\ \\
Buli kigambo mu sentensi eno wammanga giteekeko akabonero kaakyo aka NER.
Derby County eyagala okutwala Rooney , bamuwe ogwobuzannyi nga mu kiseera kye kimu nga ye mutendesi . $\to$ (Derby,B-ORG), (County,I-ORG), (eyagala,O), (okutwala,O), (Rooney,B-PER), (,,O), (bamuwe,O), (ogwobuzannyi,O), (nga,O), (mu,O), (kiseera,B-DATE), (kye,I-DATE), (kimu,I-DATE), (nga,O), (ye,O), (mutendesi,O), (.,O)
\\ \\
Buli kigambo mu sentensi eno wammanga giteekeko akabonero kaakyo aka NER.
Ono ye waffe era kampeyini ze okuziyimirizaawo tujja kwesondamu ensimbi ezinamuyamba okukuba ebipande ebipande nokukola emirimu emirara , Rose Namuli akolera ku katale ka Pepsi oluvanyuma namuwa 2 , 000 . $\to$ \textbf{(Ono,O), (ye,O), (waffe,O), (era,O), (kampeyini,O), (ze,O), (okuziyimirizaawo,O), (tujja,O), (kwesondamu,O), (ensimbi,O), (ebipande,O), (ebipande,O), (nokukola,O), (emirimu,O), (emirara,O), (Rose,B-PER), (Namuli,B-PER), (katale,I-DATE), (Pepsi,B-DATE), (2,O), (000,O), (.,O)}

        \end{minipage}}
        \caption{A 2-shot prompt in Luganda for MasakhaNER and \textbf{the output} for the \emph{Prompt} experiment}
        \end{figure}

        \begin{figure}[h]
        \noindent\fbox{\begin{minipage}[h]{\dimexpr\textwidth-80\fboxrule\relax}
        \small

        Shyira buri mutwe umutwe wamakuru hamwe ninsanganyamatsiko. Amahitamo ni 0 kuri politiki, 1 kuri siporo, 2 ku bukungu, 3 ku buzima, 4 ku myidagaduro, 5 ku mateka, 6 ku ikoranabuhanga, 7 ku bukerarugendo, 8 ku muco, 9 ku myambarire, 10 ku idini, 10 ku bidukikije , 12 ku burezi, cyangwa 13 ku mibanire \\ mu myaka imbere kanseri y’inkondo y’umura izaba amateka rwanda $\to$ 3 \\ \\ Shyira buri mutwe umutwe wamakuru hamwe ninsanganyamatsiko. Amahitamo ni 0 kuri politiki, 1 kuri siporo, 2 ku bukungu, 3 ku buzima, 4 ku myidagaduro, 5 ku mateka, 6 ku ikoranabuhanga, 7 ku bukerarugendo, 8 ku muco, 9 ku myambarire, 10 ku idini, 10 ku bidukikije , 12 ku burezi, cyangwa 13 ku mibanire \\ fardc yamaganye amakuru gahunda kugaba ibitero kunyeshyamba bufatanye nu rwanda $\to$ 0 \\ \\ Shyira buri mutwe umutwe wamakuru hamwe ninsanganyamatsiko. Amahitamo ni 0 kuri politiki, 1 kuri siporo, 2 ku bukungu, 3 ku buzima, 4 ku myidagaduro, 5 ku mateka, 6 ku ikoranabuhanga, 7 ku bukerarugendo, 8 ku muco, 9 ku myambarire, 10 ku idini, 10 ku bidukikije , 12 ku burezi, cyangwa 13 ku mibanire \\ ikipe y’ u rwanda amavubi yahesheje u rwanda agaciro itsinda benin - $\to$ \textbf{0}
        \end{minipage}}
        \caption{A 2-shot prompt in Kinyarwanda for KinNEWS and \textbf{the output} for the \emph{Prompt} experiment}
        \end{figure}

        \begin{figure}
        \noindent\fbox{\begin{minipage}{\dimexpr\textwidth-80\fboxrule\relax}
            \small

    Samar da takaitaccen labari don labarin. \\
    Saudiyya tace ba zata bari 'yan ta'adda su sa kasar a gaba ba. Sarkin ya yi sukar ne a jawabin da ya gabatar na fara azumin watan Ramadhan. Sarki Abdullah ya jaddada cewa, musulunci addini ne na hadin kai, da hakuri, yana mai bayar da tabbacin cewa ba zai taba bari 'yan ta'adda su sa kasarsa a gaba ba. Masu aikewa da labarai sun ce Saudi Arabian, na wadannan kalamai ne, sakamakon mamayar da 'yan gwagwarmayar Islama na kungiyar ISIS ke yi a makwabciyarta Iraq. $\to$ Sarki Abdullah na Saudi Arabia, ya yi suka kan abin da ya kira, fakewar da 'yan ta'adda ke yi da addini suna tafka ta'asa. \\
    
    Samar da takaitaccen labari don labarin. \\
    Faduwar farashin man ce ta sa Najeriya ta rage yawan kasafin kudinta Karamin ministan man fetur a kasar, Timipre Sylva ne ya bayyana hakan a wata sanarwa da ya fitar ranar Juma'a. Ministan ya ce daga yanzu Najeriya za ta rika hako gangar mai miliyan 1.412 a kowacce rana da miliyan 1.495 da miliyan 1.579 a tsakanin watannin Mayu zuwa Yuni, Yuli zuwa Disamba da kuma Janairun 2021 zuwa Afrilun 2022. Har wa yau, Najeriya ta bayyana fatanta cewa "idan abin ya dore" farashin man zai karu da akalla dala 15 a kan yadda yake a yanzu. Bisa kiyasin hako danyen mai na Oktoban 2018, a yanzu Najeriya na hako ganga miliyan 1.829 a kowacce rana. Wannan ragin ya biyo bayan yarjejeniyar da kasashen na OPEC tare da wasu kasashe masu arzikin fetur na rage ganga miliyan 10 na adadin da suke hakowa a kullum a yunkurinsu na farfado da darajar man. Sai dai har yanzu yarjejeniyar ba ta kullu ba sakamakon jinkirin da kasar Mexico ta yi na saka hannu a kanta. "Muna fatan idan wannan yarjejeniyar mai dumbin tarihi ta tabbata za a samu karuwar farashin mai da akalla dala 15 kan kowacce ganga a nan kusa," in ji Timipre Sylva. s$\to$ Najeriya ta bayyana cewa ta bi sahun kasashn kungiyar OPEC masu arzikin man fetur wajen rage adadin man da take hakowa a kowacce rana domin taimakawa wajen habaka farashinsa a kasuwar duniya. \\ 
    
    Samar da takaitaccen labari don labarin. \\
    Harin na ranar Litinin shine mafi muni da aka taba samu a birnin Manchester Matashin mai shekara 22 haifaffen birnin Manchester ne, dan asalin Libya kuma an yi imani iyayensa sun koma kasarsu bayan zama na tsawon shekaru. Salman Abedi ya yi makaranta a Manchester har ma ya je Jami'ar Salford da ke kusa amma ya watsar. Abokansa na tuna shi da cewa gwanin dan kwallon kafa ne amma yana tu'ammali da kwaya. Fira ministar Burtaniya Theresa May ta ce akwai yiwuwar gungun wasu daidaiku na da hannu a harin maryacen ranar Litinin da aka kai gidan raye-raye na Manchester. Ta ce Burtaniya za ta tura sojoji don tsare muhimman wurare da tarukan jama'a bayan harin kunar-bakin-wake da ya yi sanadin mutuwar mutum 22. Mrs May ce ta bayyana wannan matakin na ba-safai-ba bayan ta sanar da daga ma'aunin hukuma kan barazanar ta'addancin da Burtaniya ke fuskanta zuwa matakin la'ila-ha'ula'i. Hakan dai na nufin mai yiwuwa ne a samu karin wasu hare-hare a nan gaba. $\to$ \textbf{ada a kuma da kada a kuma da kada a kuma da kada a kuma da kada a kuma da kada a kuma da kada a kuma da kada a kuma da kada a kuma da kada a kuma da kada a kuma da kada a kuma da kada a kuma da kada a kuma da kada a kuma da kada a kuma da kada a kuma da kada a kuma da ka}
        \end{minipage}}
        \caption{A 2-shot prompt in Hausa for XL-Sum and \textbf{the output} for the \emph{Prompt} experiment}
            \label{figure11}
        \end{figure}

    \newpage
    \subsection{Examples for \emph{Translate} Experiments}
        \begin{figure}[h]
        \noindent\fbox{\begin{minipage}{\dimexpr\textwidth-80\fboxrule\relax}
            \small

        Mark each word in the following sentence with its NER. You have to plan what you will cover as the term is the best but if you don't have it you get a straw that works too $\to$ (You,O), (planning,O), (what you,O), (what,O), (it,O), (better,O), (doing,O), (good,O) , (but,O), (if,O), (don't,O), (get,O), (straw,O), (also,O), (does,O), (.,O)
\\ \\
Mark each word in the following sentence with its NER. Derby County want to take Rooney , give him a playing job at the same time as a manager $\to$ (Derby,B-ORG), (County,I-ORG), (wants,O), (take,O), (Rooney,B-PER), (,,O), (give,O) , (playful,O), (as,O), (in,O), (time,B-DATE), (of,I-DATE), (one,I-DATE), (as,O), ( he,O), (coach,O), (.,O)
\\ \\
Mark each word in the following sentence with its NER. Rose Namuli , who works at Pepsi Market , later gave him 2 , 000 to support his campaign $\to$ \textbf{(Rose,B-ORG), (Namuli,I-ORG), (works,O), (at,O), (Pepsi,I-ORG), (Market,I-ORG), (later,O), (gave,O), (him,B-PER), (2,O), (000,O), (to,O), (support,O), (his,O), (campaign,O), (.,O)}

        \end{minipage}}
        \caption{A 2-shot prompt in Luganda for MasakhaNER and \textbf{the output} for the \emph{Translate} experiment}
        \end{figure}

        \begin{figure}[h]
        \noindent\fbox{\begin{minipage}{\dimexpr\textwidth-80\fboxrule\relax}
            \small

        Give each topic a headline and topic. The options are 0 for politics, 1 for sports, 2 for economics, 3 for health, 4 for entertainment, 5 for history, 6 for technology, 7 for tourism, 8 for culture, 9 for fashion, 10 for religion, 10 for environment. 12 for education, or 13 for relationships in the years ahead of cervical cancer history $\to$ 3

        Give each topic a headline and a topic. The options are 0 for politics, 1 for sports, 2 for economics, 3 for health, 4 for entertainment, 5 for history, 6 for technology, 7 for tourism, 8 for culture, 9 for fashion, 10 for religion, 10 for environment. 12 on education, or 13 on relations, fardc has condemned the news of plans to attack rebels affiliated with Rwanda $\to$ 0
        
        Give each topic a headline and topic. The options are 0 for politics, 1 for sports, 2 for economics, 3 for health, 4 for entertainment, 5 for history, 6 for technology, 7 for tourism, 8 for culture, 9 for fashion, 10 for religion, 10 for environment, 12 on education, or 13 on relations, the Rwandan team, the tigers, made Rwanda worth the Benin team - $\to$ \textbf{0}
        \end{minipage}}
        \caption{A 2-shot prompt in Kinyarwanda for KinNEWS and the output for the \emph{Translate} experiment}
        \end{figure}

        \begin{figure}
        \noindent\fbox{\begin{minipage}{\dimexpr\textwidth-80\fboxrule\relax}
            \small

        Provide a synopsis for the story. Saudi Arabia said it will not allow terrorists to put the country first. The king made the criticism in his speech at the beginning of the month of Ramadhan. King Abdullah stressed that Islam is a religion of unity and tolerance, assuring that he will never allow terrorists to put his country first. The senders of the news said that Saudi Arabia belongs to these words, as a result of the occupation of the Islamic militants of the ISIS group in its neighbor Iraq. $\to$ King Abdullah of Saudi Arabia, criticized what he called, the hiding of terrorists by religion and committing atrocities.

        Provide a synopsis for the story. The drop in oil prices has made Nigeria reduce its budget. The Minister of Petroleum in the country, Timipre Sylva, said this in a statement he issued on Friday. The minister said that from now on, Nigeria will produce 1.412 million barrels of oil per day and 1.495 million and 1.579 million barrels between the months of May to June, July to December and January 2021 to April 2022. Until today, Nigeria expressed its hope that ``if it lasts"; the price of oil will increase by at least 15 dollars on the current level. According to the crude oil production estimate for October 2018, Nigeria is currently producing 1.829 million barrels per day. This reduction follows the agreement of the OPEC countries with other oil rich countries to reduce 10 million barrels of the amount they produce every day in their efforts to revive the value of oil. However, the agreement has not yet been finalized due to Mexico's delay in signing it. ``We hope that if this historic agreement is confirmed, there will be an increase in the price of oil by at least 15 dollars per barrel in the near future,"; said Timipre Sylva. $\to$ Nigeria has announced that it has joined the OPEC group of oil rich countries in reducing the amount of oil it produces every day to help increase its price in the world market.
        
        Provide a synopsis for the story. Monday's attack was the worst ever in Manchester. Salman Abedi went to school in Manchester and even went to the nearby University of Salford but dropped out. His friends remember him as a good football player but he was a bit of a jerk. British Prime Minister Theresa May said there is a possibility that a group of individuals were involved in Monday's attack on a Manchester nightclub. She said the UK would send troops to secure important places and public gatherings after the arson attack that left 22 dead. Mrs May made the rare move after announcing the lifting of the The UK's terror threat rating has been raised to moderate. This means that there may be more attacks in the future. $\to$ \textbf{The British Prime Minister said that the UK will send troops to secure important places and public gatherings after the arson attack that left 22 dead. Mrs May made the rare move after announcing the lifting of the The UK's terror threat rating has been raised to moderate. This means that there may be more attacks in the future. \\ 
        Provide a synopsis for the story. The United States has announced that it will send a delegation to the G7 summit in Italy. The White House said that the delegation will be led by the US Secretary of State, Rex Tillerson. The White House said that the delegation will be led by the US Secretary of State, Rex Tillerson. The White House said that the delegation will be led by the US Secretary of State, Rex Tillerson. The White House said that the delegation will be led by the US Secretary of State, Rex Tillerson. The White House said that the delegation will be led by the US Secretary of State, Rex Tillerson. The White House said that the delegation will be led by the US Secretary of State, Rex Tillerson. The White House said that the delegation will be led}
        \end{minipage}}
        \caption{A 2-shot prompt in Hausa for XL-Sum and the output for the \emph{Translate} experiment}
        \label{figure14}
        \end{figure}

    \newpage
    \subsection{Examples for \emph{Language-adaptive fine-tuning} Experiments}
        \begin{figure}[h]
        \noindent\fbox{\begin{minipage}{\dimexpr\textwidth-80\fboxrule\relax}
            \small

       Buli kigambo mu sentensi eno wammanga giteekeko akabonero kaakyo aka NER.
Olina okutegeka byonoobikkisa ngessanja lye lisinga okukola obulungi naye bwoba tolirina ofuna essubi nalyo likola . $\to$ (Olina,O), (okutegeka,O), (byonoobikkisa,O), (ngessanja,O), (lye,O), (lisinga,O), (okukola,O), (obulungi,O), (naye,O), (bwoba,O), (tolirina,O), (ofuna,O), (essubi,O), (nalyo,O), (likola,O), (.,O)
\\ \\
Buli kigambo mu sentensi eno wammanga giteekeko akabonero kaakyo aka NER.
Derby County eyagala okutwala Rooney , bamuwe ogwobuzannyi nga mu kiseera kye kimu nga ye mutendesi . $\to$ (Derby,B-ORG), (County,I-ORG), (eyagala,O), (okutwala,O), (Rooney,B-PER), (,,O), (bamuwe,O), (ogwobuzannyi,O), (nga,O), (mu,O), (kiseera,B-DATE), (kye,I-DATE), (kimu,I-DATE), (nga,O), (ye,O), (mutendesi,O), (.,O)
\\ \\
Buli kigambo mu sentensi eno wammanga giteekeko akabonero kaakyo aka NER.
Ono ye waffe era kampeyini ze okuziyimirizaawo tujja kwesondamu ensimbi ezinamuyamba okukuba ebipande ebipande nokukola emirimu emirara , Rose Namuli akolera ku katale ka Pepsi oluvanyuma namuwa 2 , 000 . $\to$  \textbf{(Ono,O), (era,O), (kampeyini,O), (ze,O), (okuziyimirizaawo,O), (tujja,O), (kwesondamu,O), (ensimbi,O), (ezinamuyamba,O), (okukuba,O), (ebipande,O), (ebipande,O), (nokukola,O), (emirimu,O), (emirara,O), (.,O)}

        \end{minipage}}
        \caption{A 2-shot prompt in Luganda for MasakhaNER and the output for the \emph{language-adaptive fine-tuning} experiment}
        \end{figure}

        \begin{figure}[h]
        \noindent\fbox{\begin{minipage}{\dimexpr\textwidth-80\fboxrule\relax}
            \small

        Shyira buri mutwe umutwe wamakuru hamwe ninsanganyamatsiko. Amahitamo ni 0 kuri politiki, 1 kuri siporo, 2 ku bukungu, 3 ku buzima, 4 ku myidagaduro, 5 ku mateka, 6 ku ikoranabuhanga, 7 ku bukerarugendo, 8 ku muco, 9 ku myambarire, 10 ku idini, 10 ku bidukikije , 12 ku burezi, cyangwa 13 ku mibanire
        mu myaka imbere kanseri y’inkondo y’umura izaba amateka rwanda $\to$ 3
        
        Shyira buri mutwe umutwe wamakuru hamwe ninsanganyamatsiko. Amahitamo ni 0 kuri politiki, 1 kuri siporo, 2 ku bukungu, 3 ku buzima, 4 ku myidagaduro, 5 ku mateka, 6 ku ikoranabuhanga, 7 ku bukerarugendo, 8 ku muco, 9 ku myambarire, 10 ku idini, 10 ku bidukikije , 12 ku burezi, cyangwa 13 ku mibanire
        fardc yamaganye amakuru gahunda kugaba ibitero kunyeshyamba bufatanye nu rwanda $\to$ 0
        
        Shyira buri mutwe umutwe wamakuru hamwe ninsanganyamatsiko. Amahitamo ni 0 kuri politiki, 1 kuri siporo, 2 ku bukungu, 3 ku buzima, 4 ku myidagaduro, 5 ku mateka, 6 ku ikoranabuhanga, 7 ku bukerarugendo, 8 ku muco, 9 ku myambarire, 10 ku idini, 10 ku bidukikije , 12 ku burezi, cyangwa 13 ku mibanire
        ikipe y’ u rwanda amavubi yahesheje u rwanda agaciro itsinda benin - $\to$ \textbf{0}
        \end{minipage}}
        \caption{A 2-shot prompt in Kinyarwanda for KinNEWS and the output for the \emph{language-adaptive fine-tuning} experiment}
        \end{figure}

        \newpage
        \begin{figure}[h]
        \noindent\fbox{\begin{minipage}{\dimexpr\textwidth-80\fboxrule\relax}
            \small

        Samar da takaitaccen labari don labarin.
        Saudiyya tace ba zata bari 'yan ta'adda su sa kasar a gaba ba. Sarkin ya yi sukar ne a jawabin da ya gabatar na fara azumin watan Ramadhan. Sarki Abdullah ya jaddada cewa, musulunci addini ne na hadin kai, da hakuri, yana mai bayar da tabbacin cewa ba zai taba bari 'yan ta'adda su sa kasarsa a gaba ba. Masu aikewa da labarai sun ce Saudi Arabian, na wadannan kalamai ne, sakamakon mamayar da 'yan gwagwarmayar Islama na kungiyar ISIS ke yi a makwabciyarta Iraq. $\to$ Sarki Abdullah na Saudi Arabia, ya yi suka kan abin da ya kira, fakewar da 'yan ta'adda ke yi da addini suna tafka ta'asa.
        
        Samar da takaitaccen labari don labarin.
        Faduwar farashin man ce ta sa Najeriya ta rage yawan kasafin kudinta Karamin ministan man fetur a kasar, Timipre Sylva ne ya bayyana hakan a wata sanarwa da ya fitar ranar Juma'a. Ministan ya ce daga yanzu Najeriya za ta rika hako gangar mai miliyan 1.412 a kowacce rana da miliyan 1.495 da miliyan 1.579 a tsakanin watannin Mayu zuwa Yuni, Yuli zuwa Disamba da kuma Janairun 2021 zuwa Afrilun 2022. Har wa yau, Najeriya ta bayyana fatanta cewa "idan abin ya dore" farashin man zai karu da akalla dala 15 a kan yadda yake a yanzu. Bisa kiyasin hako danyen mai na Oktoban 2018, a yanzu Najeriya na hako ganga miliyan 1.829 a kowacce rana. Wannan ragin ya biyo bayan yarjejeniyar da kasashen na OPEC tare da wasu kasashe masu arzikin fetur na rage ganga miliyan 10 na adadin da suke hakowa a kullum a yunkurinsu na farfado da darajar man. Sai dai har yanzu yarjejeniyar ba ta kullu ba sakamakon jinkirin da kasar Mexico ta yi na saka hannu a kanta. "Muna fatan idan wannan yarjejeniyar mai dumbin tarihi ta tabbata za a samu karuwar farashin mai da akalla dala 15 kan kowacce ganga a nan kusa," in ji Timipre Sylva. $\to$ Najeriya ta bayyana cewa ta bi sahun kasashn kungiyar OPEC masu arzikin man fetur wajen rage adadin man da take hakowa a kowacce rana domin taimakawa wajen habaka farashinsa a kasuwar duniya.
        
        Samar da takaitaccen labari don labarin.
su        Harin na ranar Litinin shine mafi muni da aka taba samu a birnin Manchester Matashin mai shekara 22 haifaffen birnin Manchester ne, dan asalin Libya kuma an yi imani iyayensa sun koma kasarsu bayan zama na tsawon shekaru. Salman Abedi ya yi makaranta a Manchester har ma ya je Jami'ar Salford da ke kusa amma ya watsar. Abokansa na tuna shi da cewa gwanin dan kwallon kafa ne amma yana tu'ammali da kwaya. Fira ministar Burtaniya Theresa May ta ce akwai yiwuwar gungun wasu daidaiku na da hannu a harin maryacen ranar Litinin da aka kai gidan raye-raye na Manchester. Ta ce Burtaniya za ta tura sojoji don tsare muhimman wurare da tarukan jama'a bayan harin kunar-bakin-wake da ya yi sanadin mutuwar mutum 22. Mrs May ce ta bayyana wannan matakin na ba-safai-ba bayan ta sanar da daga ma'aunin hukuma kan barazanar ta'addancin da Burtaniya ke fuskanta zuwa matakin la'ila-ha'ula'i. Hakan dai na nufin mai yiwuwa ne a samu karin wasu hare-hare a nan gaba. $\to$ \textbf{dan kunar-bakin-wake Salman Abedi, ya kai harin da ya hallaka mutum 22 a gidan raye-raye na Manchester Arena da ke birnin Manchester a Ingila. Mutane da dama ne kuma suka jikkata a harin da aka kai da misalin karfe 10:30 na dare agogon GMT, yayin da mawakiya Ariana Grande ke kammala wasan da ta yi a gidan raye-rayen. 'Yan sanda sun ce mutum 64 ne suka samu raunuka, wasu cikin mawuyacin hali. An rufe dukkan makarantun da ke yankin a lokacin harin, kuma an umarci dukkan dalibai su koma gida. An kuma rufe dukkan tashoshin jiragen kasa da ke yankin. Fira ministar Burtaniya Theresa May ta ce an dauki matakin ne don tabbatar da tsaron}
        \end{minipage}}
        \caption{A 2-shot prompt in Hausa for XL-Sum and the output for the \emph{language-adaptive fine-tuning} experiment}
        \label{figure17}
        \end{figure}

    \newpage
    \subsection{Reproducibility}
    \label{reproducibility}
    We use the Google Cloud Translate API for translation. The total cost for translating the few-shot prompts, the test sets, and the generated output for all experiments costs just under \$600. 
    
    We use a cluster of 4 NVIDIA A100 GPUs with 80 GB of RAM each. The LAFT Hausa model took approximately 99 hours to train, the LAFT Kinyarwanda model took approximately 4.3 hours to train, and the LAFT Luganda model took approximately 0.01 hours to train. In total, LAFT took about 412 GPU hours, which approximately costs \$2,700 after applying current market prices for training costs only. 
    \subsection{Hyperparameters}
    \label{hyperparams}
    We perform language-adaptive fine-tuning for each language and utilize the following hyperparameters: $3$ epochs, warmup ratio of $0.03$, no weight decay, and a learning rate of $2e^{-5}$

    \subsection{LAFT Training Graphs}
    \begin{figure}[h]
        \centering
        \includegraphics[width=0.6\textwidth]{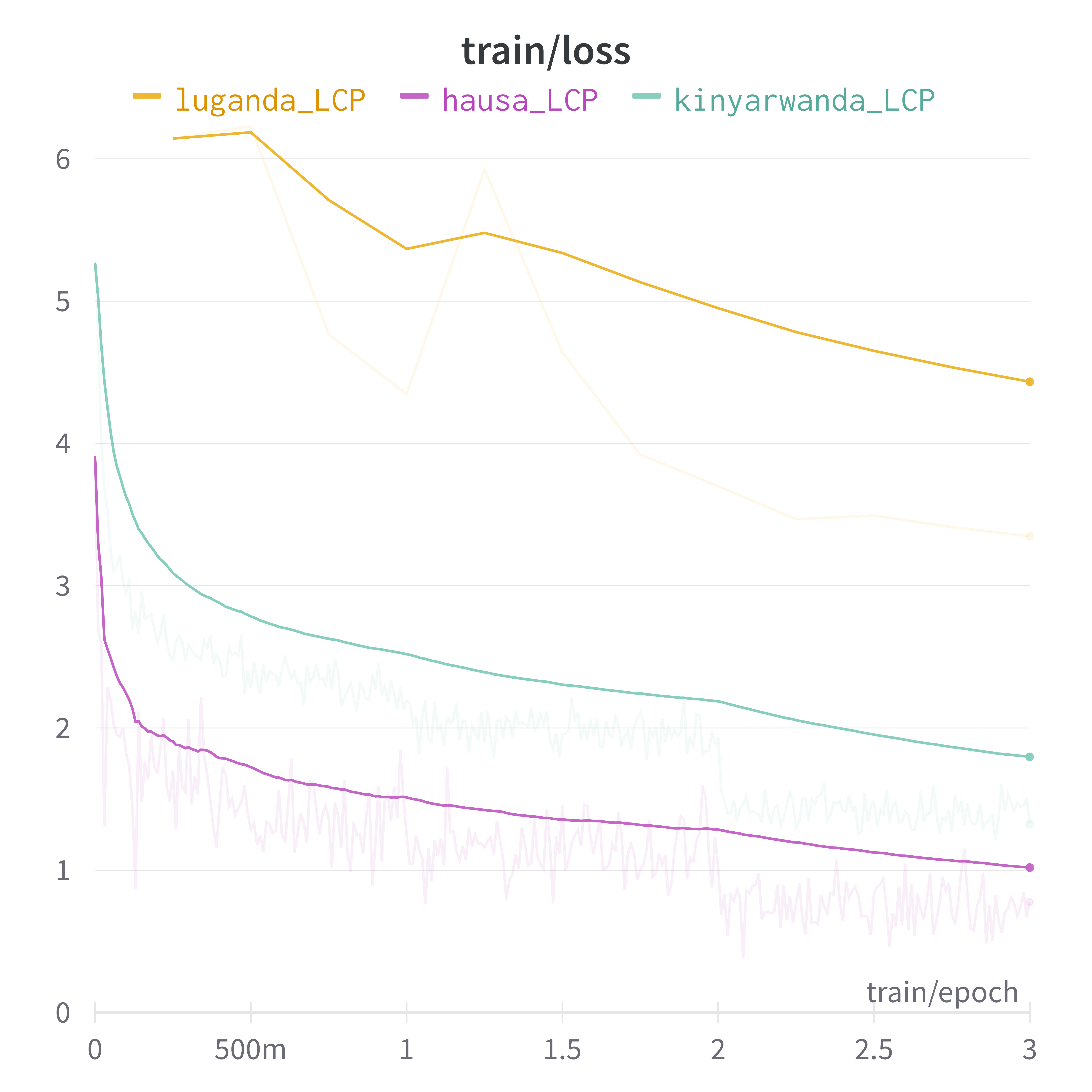}
        \caption{LAFT Train Loss Over Epochs}
    \end{figure}
    \begin{figure}[h]
        \centering
        \includegraphics[width=0.6\textwidth]{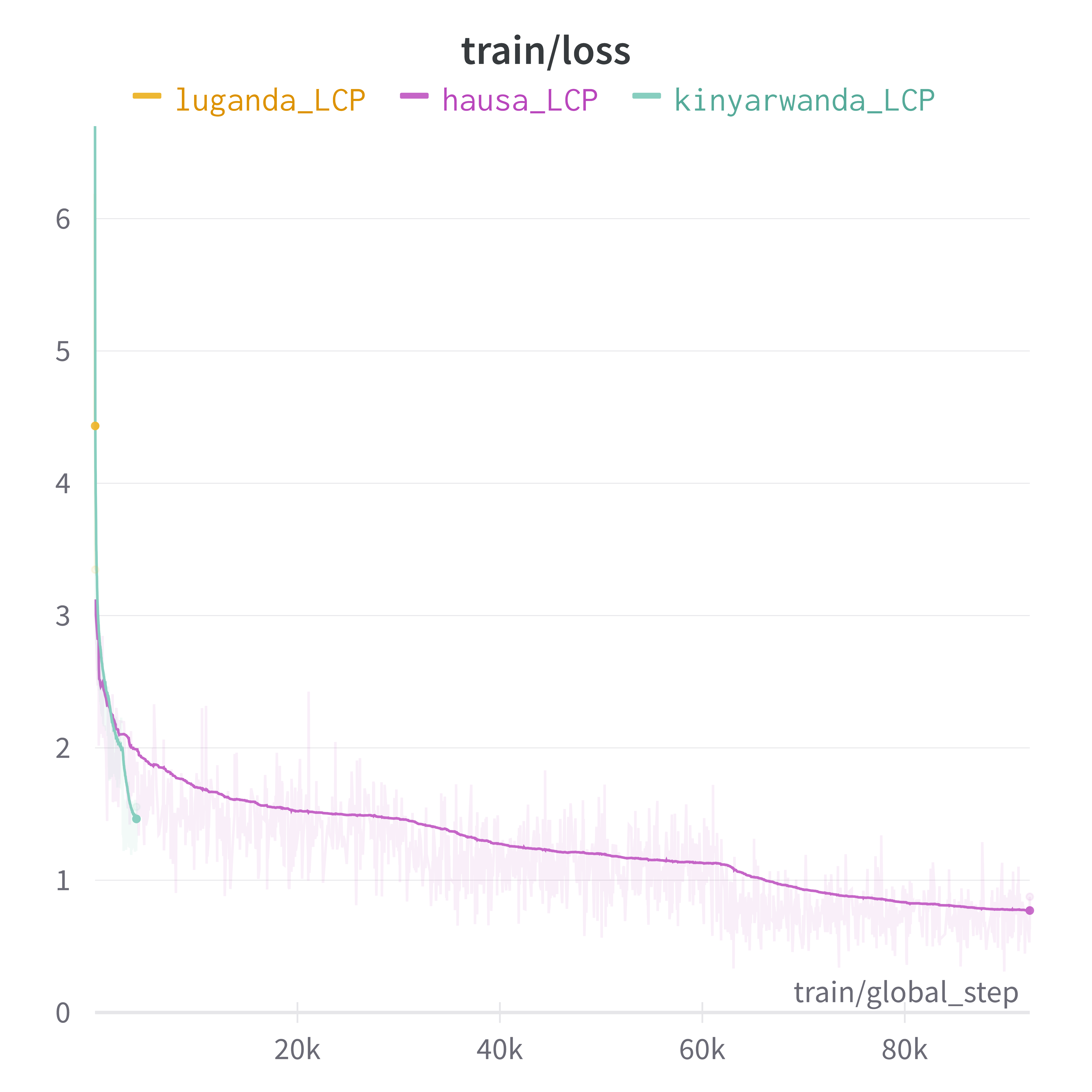}
        \caption{LAFT Train Loss Over Training Steps}
    \end{figure}

    \clearpage
    
    \subsection{Performance of LLaMa vs. GPT-3 for few-shot in-context learning} \label{gpt3llama}

    \begin{table*}[h]
    \centering
    \begin{tabular}{c|c||c|c|c|c|c}
    \hline \hline Model & \# Params. & 0-shot & 1-shot & 5-shot & 64-shot & Reported By \\ \hline
    GPT-3 Small & 125 M & 0.64 & 1.19 & -- & 1.72 &   \\
    GPT-3 Medium & 350 M & 1.75  & 3.07 & -- & 4.46 &   \\
    GPT-3 Large & 760 M & 2.71 & 4.79 & -- & 7.89 &   \\
    GPT-3 XL & 1.3 B & 4.40 & 5.43 & -- & 9.72 & \cite{gpt3}   \\
    GPT-3 2.7B & 2.7 B & 6.01 & 8.73  & -- & 13.2 &  \\
    GPT-3 6.7B & 6.7 B & 5.79 & 9.78  & -- & 17.0 &   \\
    GPT-3 13B & 13 B & 7.84 & 13.7 & -- & 21.0 &   \\
    GPT-3 175B & 175 B  & 14.6 & 23.0 & -- & 29.9 &   \\ \hline
    LLaMa 7B & 7 B & 16.8 & 18.7 & 22.0 & 26.1 &   \\
    LLaMa 13B & 13 B & 20.1 & 23.4 & 28.1 & 31.9 &  \cite{touvron2023llama} \\
    LLaMa 33B & 33 B & 24.9 & 28.3 & 32.9 & 36.0 &   \\
    LLaMa 65B & 65 B & 23.8 & 31.0 & 35.0 & 39.9 &   \\ \hline

    \hline
    \end{tabular}
    \caption{LLaMa and GPT-3 Prompting Accuracy on NaturalQuestions test set with the Exact Answer Match metric}

    \end{table*}

    \newpage
    \begin{table*}[h]
    \centering
    \begin{tabular}{c|c||c|c|c|c|c}
    \hline \hline Model & \# Params. & 0-shot & 1-shot & 5-shot & 64-shot & Reported By \\ \hline
    GPT-3 Small & 125 M  & 4.15 & 4.19 & -- & 6.96 &   \\
    GPT-3 Medium & 350 M & 7.61  & 12.9 & -- & 16.3 &   \\
    GPT-3 Large & 760 M & 14.0 & 20.5 & -- & 26.5 &   \\
    GPT-3 XL & 1.3 B & 19.7 & 26.5 & -- & 32.1& \cite{gpt3}   \\
    GPT-3 2.7B & 2.7 B & 31.3 & 35.9  & -- & 42.3 &  \\
    GPT-3 6.7B & 6.7 B & 38.7 & 44.4  & -- & 51.6 &   \\
    GPT-3 13B & 13 B & 41.8 & 51.3 & -- & 57.5 &   \\
    GPT-3 175B & 175 B & 64.3 & 68.0 & -- & 71.2 &   \\ \hline
    LLaMa 7B & 7 B & 50.0 & 53.4 & 56.3 & 57.6 &   \\
    LLaMa 13B & 13 B & 56.6 & 60.5 & 63.1 & 64.0 &  \cite{touvron2023llama} \\
    LLaMa 33B & 33 B & 65.1 & 67.9 & 69.9 & 70.4 &   \\
    LLaMa 65B & 65 B & 68.2 & 71.6 & 72.6 & 73.0 &   \\ \hline

    \hline
    \end{tabular}
    \caption{LLaMa and GPT-3 Prompting Accuracy on TriviaQA dev set with the Exact Answer Match metric}

    \end{table*}

\end{document}